\documentclass[11pt]{article}

\usepackage{color}
\usepackage{algorithm, algpseudocode}
\usepackage{mathrsfs}
\usepackage{dsfont}
\usepackage{lmodern}
\usepackage{array}
\usepackage{bbm}
\usepackage{amsmath}
\usepackage{amssymb}
\usepackage[mathscr]{eucal}
\usepackage{graphicx}
\usepackage{mathrsfs}
\usepackage{psfrag}
\usepackage{color}
\usepackage{here}
\usepackage[margin=0.95in]{geometry}
\usepackage{wasysym}
\usepackage{enumitem}
\usepackage{xcolor}
\usepackage{caption}
\usepackage{subcaption}
\usepackage{amsthm}
\usepackage{lipsum}
\usepackage{hyperref}

\newcommand{\Exp}{\mathds{E}}

\newcommand{\Prob}{\mathds{P}}
\newcommand{\Real}{\mathds{R}}
\newcommand{\Nat}{\mathbb{N}}

\newcommand{\Yc}{\mathcal{Y}}

\newcommand{\Ac}{\mathcal{A}}

\title{Bootstrapped Thompson Sampling and Deep Exploration}

\author{
Ian Osband
\and
Benjamin Van Roy
}

\begin{document}
\maketitle


\section*{Abstract}

This technical note presents a new approach to carrying out the kind of exploration achieved by Thompson sampling,
but without explicitly maintaining or sampling from posterior distributions.  The approach is based on a bootstrap
technique that uses a combination of observed and artificially generated data.  The latter serves to induce a prior
distribution which, as we will demonstrate, is critical to effective exploration.  We explain how the approach
can be applied to multi-armed bandit and reinforcement learning problems and how it relates to Thompson sampling.  The approach is
particularly well-suited for contexts in which exploration is coupled with deep learning, since in these settings,
maintaining or generating samples from a posterior distribution becomes computationally infeasible.


\section{Introduction}

To perform well in a sequential decision task while learning about its environment, an agent must balance between exploitation, making good decisions given available
data, and exploration, taking actions that may help to improve the quality of future decisions.
Perhaps the most principled approach is to compute a Bayes optimal solution, which optimizes the long-run expected rewards given prior beliefs.
Although conceptually simple, this approach is computationally intractable for all but the simplest of problems.  As such, engineers
typically turn to tractable heuristic exploration strategies.

Upper-confidence bound approaches offer one popular class of exploration heuristics that come with performance guarantees.
Such approaches assign to poorly-understood actions high but statistically plausible values, effectively allocating an optimism bonus to
incentivize exploration of each such action.  For a broad class of problems, if optimism bonuses are well-designed, upper-confidence
bound algorithms enjoy optimal learning rates.  However, designing, tuning, and applying such algorithms can be challenging or intractable,
and as such, upper-confidence bound algorithms applied in practice often suffer poor empirical performance.

Another popular heuristic, which on the surface appears unrelated, is called Thompson sampling or probability matching.
In this algorithm the agent maintains a posterior distribution of beliefs and, ain each time period, samples an action randomly according to the probability that
it is optimal.
Although this algorithm is one of the oldest recorded exploration heuristics, it received relatively little attention until recently when
its strong empirical performance was noted, and a host of analytic guarantees followed.
It turns out that there is a deep connection between Thompson sampling and optimistic algorithms; in particular, as shown in
\cite{russo2014}, Thompson sampling can be viewed as a randomized approximation that approaches the performance of
a well-designed and well-tuned upper confidence bound algorithm.

Almost all of the literature on Thompson sampling takes the ability to sample from a posterior distribution as given.
For many commonly used distributions, this is served through conjugate updates or Markov chain Monte Carlo methods.
However, such methods do not adequately accommodate contexts in which models are nonlinearly parameterized
in potentially complex ways, as is the case in deep learning.  In this paper we introduce an alternative approach to
tractably attaining the behavior of Thompson sampling in a manner that accommodates such nonlinearly parameterized
models and, in fact, may offer advantages more broadly when it comes to efficiently implementing and applying Thompson sampling.

The approach we propose is based on a bootstrap technique that uses a combination of observed and artificially generated data.
The idea of using the bootstrap to approximate a posterior distribution is not new,
and has been noted from inception of the bootstrap concept.
Further, the application of the bootstrap \cite{DBLP:journals/corr/EcklesK14} and other related sub-sampling approaches \cite{baransi2014sub} to
approximate Thompson sampling is not new.
However, we show that these existing approaches fail to ensure sufficient exploration for effective performance in sequential decision problems.
As we will demonstrate, the way in which we generate and use artificial data is critical.
The approach is particularly well-suited for contexts in which exploration is coupled with deep learning, since in such settings, maintaining or generating
samples from a posterior distribution becomes computationally infeasible.  Further, our approach is parallelizable and as such
scales well to massive complex problems.  We explain how the approach
can be applied to multi-armed bandit and reinforcement learning problems and how it relates to Thompson sampling.


\section{Priors, posteriors, and the bootstrap}

The term {\it bootstrap} refers to a class of methods for nonparametric estimation from data-driven simulation \cite{efron1979bootstrap}.
In essence, the bootstrap uses the empirical distribution of a sampled dataset as an estimate of the population statistic.
Algorithm \ref{alg:Bootstrap} provides pseudocode for what is perhaps the most common form of bootstrap \cite{efron1979bootstrap}.
We use $\mathcal{P}(\mathcal{X})$ to denote the set of probability measures over a set $\mathcal{X}$.

\begin{algorithm}[H]
\caption{\protect\\ Bootstrap}
\label{alg:Bootstrap}
{\textbf{Input:} Data $x_1,..,x_N \in \mathcal{X}$, function $\phi: \mathcal{P}(\mathcal{X}) \mapsto \mathcal{Y}$, $K \in \Nat$} \\
{\textbf{Output:} Probability measure $\hat{P} \in \mathcal{P}(\mathcal{Y})$}
{\small
\begin{algorithmic}[1]
    \For{$k=1,..,K$}
        \State{sample data $x^k_1,..,x^k_N$ from $\{x_1,\ldots,x_N\}$ uniformly with replacement}
        	\State{for all $dx \subseteq \mathcal{X}$, let $\hat{P}_k(dx) = \sum_{n=1}^N {\bf 1}(x^k_n \in dx) / N$}
        \State{compute $y_k = \phi(\hat{P}_k)$}
    \EndFor
    \State{For all $dy \in \mathcal{Y}$, let $\hat{P}(dy) = \sum_{k=1}^K {\bf 1}(y_k \in d y) / K$}
\end{algorithmic}
}
\end{algorithm}
\vspace{-3mm}

This procedure allows estimates the distribution of any unknown parameter in a non-parametric manner.  As described with
a function $\phi$ of probability measures as input, the algorithm is somewhat abstract.  However, it easily specializes
to familiar concrete versions.  For example, suppose $\phi$ is the expectation operator.  Then, each sample $y_k$
is the mean of a sample of the data set, and $\hat{P}$ is the relative frequency measure of these sample means.

The output $\hat{P}$ is reminiscent of a Bayesian posterior with an extremely weak prior.
In fact, with a small modification, Algorithm \ref{alg:Bootstrap} becomes Algorithm \ref{alg:BayesBootstrap}, the so-called
Bayesian bootstrap, for which the distribution produced can be interpreted as a posterior based on the data
and a degenerate Dirichlet prior \cite{rubin1981bayesian}.

\begin{algorithm}[H]
\caption{\protect\\  BayesBootstrap}
\label{alg:BayesBootstrap}
{\textbf{Input:} Data $x_1,..,x_N \in \mathcal{X}$, function $\phi: \mathcal{P}(\mathcal{X}) \mapsto \mathcal{Y}$, $K \in \Nat$} \\
{\textbf{Output:} Probability measure $\hat{P} \in \mathcal{P}(\mathcal{Y})$}
{\small
\begin{algorithmic}[1]
    \For{$k=1,..,K$}
    	\State{sample $w^k_1, \ldots, w^k_N \sim {\rm Exp}(1)$}
	\State{for all $dx \subseteq \mathcal{X}$, let $\hat{P}_k(dx) = \sum_{n=1}^N w^k_n {\bf 1}(x_n \in dx) / \sum_{n=1}^N w^k_n$}
        \State{compute $y_k = \phi(\hat{P}_k)$}
    \EndFor
    \State{For all $dy \in \mathcal{Y}$, let $\hat{P}(dy) = \sum_{k=1}^K {\bf 1}(y_k \in d y) / K$}
\end{algorithmic}
}
\end{algorithm}

With the bootstrap approaches we have described, the support of distributions $\hat{P}_k$ is restricted to the dataset
$\{x_1,..,x_N\}$.
We will show that in sequential decision problems this poses a significant problem.
To address the problem, we propose a simple extension to the bootstrap.
In particular, we augment the dataset $\{x_1,\ldots, x_N\}$ with artificially generated samples
$\{x_{N+1},..,x_{N+M}\}$ and apply the bootstrap to the combined dataset.  The artificially generated data can be viewed
as inducing a prior distribution.
In fact, if $\mathcal{X}$ is finite and $\{x_{N+1},..,x_{N+M}\} = \mathcal{X}$, then as $K$ grows,
the distribution $\hat{P}$ produced by the Bayesian  bootstrap converges to the posterior distribution conditioned on $\{x_1,\ldots, x_N\}$,
given a uniform Dirichlet prior.

When the set $\mathcal{X}$  is large or infinite, the idea of generating one artificial sample per element does not scale gracefully.
If we require one prior observation for every single possible value then the observed data $\{x_1,\ldots, x_N\}$ will bear little influence
on the distribution $\hat{P}$ relative to the artificial data $\{x_{N+1},..,x_{N+M}\}$.  To address this, for a selected value of $M$, we sample the $M$
artificial data points $x_{N+1},..,x_{N+M}$ from a ``prior'' distribution $P_0$.
The important thing here is that the relative strength $M / N$ of the induced prior can be controlled in an explicit manner.
Similarly, depending on the choice of the prior sampling distribution $P_0$ the posterior is no longer restricted to finite support.
In many ways this extension corresponds to using a Dirichlet process prior with generator $P_0$.

This augmented bootstrap procedure is especially promising for nonlinear functions such as deep neural networks, where sampling from
a posterior is typically intractable.
In fact, given $\emph{any}$ method of training a neural network on a dataset of size $N$, we can generate an approximate posterior
through $K$ bootstrapped versions of the neural network.  In its most naive implementation this increases the computational cost by a factor of $K$.
However, this approach is parallelizable and therefore scalable with compute power.
In addition, it may be possible to significantly reduce the computational cost of this bootstrap procedure by sharing some of the lower-level features between bootstrap sampled networks or growing them out in a tree structure.
This could even be implemented on a single chip through a specially constructed dropout mask for each bootstrap sample.
With a deep neural network this could provide significant savings.

\section{Multi-armed bandit}

Consider a problem in which an agent sequentially chooses actions $(A_t: t \in \Nat)$ from an action set $\mathcal{A}$ and observes corresponding outcomes $(Y_{t,A_t}: t \in \Nat)$.
There is a random outcome $Y_{t,a} \in \mathcal{Y}$ associated with each $a \in \mathcal{A}$ and time $t \in \Nat$.
For each random outcome the agent receives a reward $R(Y_{t,a})$ where $R: \Yc \mapsto \Real$ is a known function.
The ``true outcome distribution'' $p^*$ is itself drawn from a family of distributions $\mathcal{P}$.
We assume that, conditioned on $p^*$, $(Y_t: t \in \Nat)$ is an iid sequence with each element $Y_t$ distributed according to $p^*$.
Let $p^*_a$ be the marginal distribution corresponding to $Y_{t,a}$.
The $T$ -period regret of the sequence of actions $A_1,\ldots,A_T$ is the random variable
$${\rm Regret}(T,p^*) = \sum_{t=1}^T
    \Exp\left[\max_a R(Y_{t,a}) - R(Y_{t,A_t}) \ | \ p^*\right].$$
The Bayesian regret to time $T$ is defined by ${\rm BayesRegret}(T) = \Exp[{\rm Regret}(T, p^*)]$, where the expectation is taken with respect to the prior distribution over $p^*$.

We take all random variables to be defined with respect to a probability space $(\Omega,\mathcal{F},\Prob)$.
We will denote by $H_t$ the history of observations $(A_1, Y_{1,A_1}, .., A_{t-1}, Y_{t-1,A_{t-1}})$ realized prior to time $t$.
Each action $A_t$ is selected based only on $H_t$ and possibly some external source of randomness.  To represent this external source,
we introduce a sequence of iid random variables $(U_t: t \in \Nat)$.  Each action $A_t$ is measurable with respect to the sigma-algebra
generated by $(H_t,U_t)$.

The objective is to choose actions in a manner that minimizes Bayesian regret.  For this purpose, it is useful to think of actions as being selected by a
randomized policy $\pi = (\pi_t: t \in \Nat)$, where each $\pi_t$ is a distribution over actions and is measurable with respect to
the sigma-algebra generated by $H_t$.  An action $A_t$ is chosen at time $t$ by randomizing according to $\pi_t(\cdot) = \Prob(A_t \in \cdot | H_t)$.
Our bootstrapped Thompson sampling algorithm, presented as Algorithm \ref{alg:Boot Thompson}, serves as such a policy.
The algorithm uses a bootstrap algorithm, like Algorithms \ref{alg:Bootstrap} or \ref{alg:BayesBootstrap} as a subroutine.
Note that the sequence of action-observation pairs is not iid, though bootstrap algorithms effectively treat data passed to them as iid.
The function $\Exp[p^* | H_{t+M} = \cdot]$ passed to the bootstrap algorithm maps a specified history of $t+M$ action-observation pairs to a probability distribution
over reward vectors.  The resulting probability distribution can be thought of as a model fit to the data provided in the history.
Note that the algorithm takes as input a distribution $\tilde{P}$ from which artificial samples are drawn.  This can be thought of as a subroutine
that generates $M$ action-observation pairs.  As a special case, this subroutine could generate $M$ deterministic pairs.  There can be advantages,
though, to using a stochastic sampling routine, especially when the space of action-observation pairs is large and we do not want to
impose too strong a prior.

\begin{algorithm}[H]
\caption{\protect\\ BootstrapThompson}
\label{alg:Boot Thompson}
{\textbf{Input:} Bootstrap algorithm $B$, artificial history length $M$ and sampling distribution $\tilde{P}$}
{\small
\begin{algorithmic}[1]
\State $H_1 = ()$
    \For{$t=1,2,..$}
    \State{Sample artificial history $\tilde{H} = ((\tilde{A}_1, \tilde{Y}_1), \ldots, (\tilde{A}_M, \tilde{Y}_M)) \sim \tilde{P}$}
    \State{Bootstrap sample $\hat{P} =
    \text{B}(\tilde{H} \cup H_t, \ \Exp[p^* | H_{t + M} = \cdot], \ K=1)$}
    \State{Sample $\hat{p} \sim \hat{P}$}
    \State{Select $A_t \in \arg \max_a \Exp[R(Y_{t,a}) | \hat{p}]$}
    \State{Observe $Y_{t,A_t}$}
    \State{Update $H_{t+1} = H_t \cup (A_t, Y_{t,A_t})$}
    \EndFor
\end{algorithmic}
}
\end{algorithm}

This algorithm is similar to Thompson sampling, though the posterior sampling step has been replaced by a single bootstrap sample.
As we will establish in Section \ref{sec: Regret}, for several multi-armed bandit problems BootstrapThompson with appropriate artificial data is equivalent to Thompson sampling.

One drawback of Algorithm \ref{alg:Boot Thompson} is that the computational cost per timestep grows with the amount of data $H_t$.
For applications at large scale this will be prohibitive.
Fortunately there is an effective method to approximate these bootstrap samples in an online manner at constant computational cost \cite{DBLP:journals/corr/EcklesK14}.
Instead of generating a new bootstrap sample every step we can approximate the bootstrap bootstrap distribution by training $D \in \Nat$ online bootstrap models in parallel and then sampling between them uniformly.
In its most naive implementation this parrallel bootstrap will have a computational cost per timestep $D$ times larger than a greedy algorithm.
However, for specific function classes such as neural networks it may be possible to share some computation between models and provide significant savings.

\subsection{Simulation results}

We now examine a simple problem instance designed to demonstrate the need for an artificial history to incentivize efficient exploration in BootstrapThompson.
The action space $\Ac = \{1, 2\}$, outcomes $\Yc = [0,1]$ and rewards $R(y) = y$.
We fix $0 < \epsilon \ll 1$ and describe the true underlying distribution in terms of the Dirac delta function $\delta_{x}(y)$ which assigns all probability mass to $x$:
\begin{equation*}
 p^*_a(y) = \begin{cases}
            \delta_\epsilon(y) &\text{if $a = 1$}\\
            (1- 2\epsilon) \delta_0(y) + 2 \epsilon \delta_1(y) &\text{if $a = 2$}
        \end{cases}.
\end{equation*}
The optimal policy is to pick $a=2$ at every timestep, since this has an expected reward of $2 \epsilon$ instead of just $\epsilon$.
However, with probability at least $1 - 2\epsilon$, BootstrapThompson without artificial history ($M=0$) will \emph{never} learn the optimal policy.

To see why this is the case note that BootstrapThompson without artificial history must begin by sampling each arm once.
In our system this means that with probability $1-2\epsilon$ the agent will receive a reward of $\epsilon$ from arm one and $0$ from arm two.
Given this history, the algorithm will prefer to choose arm one for \emph{all} subsequent timesteps, since its bootstrap estimates will always put all their mass on $\epsilon$ and $0$ respectively.
However, we show that this failing can easily be remedied by the inclusion of some artificial history.

\begin{figure}[h!]
  \centering
    \includegraphics[width=\textwidth]{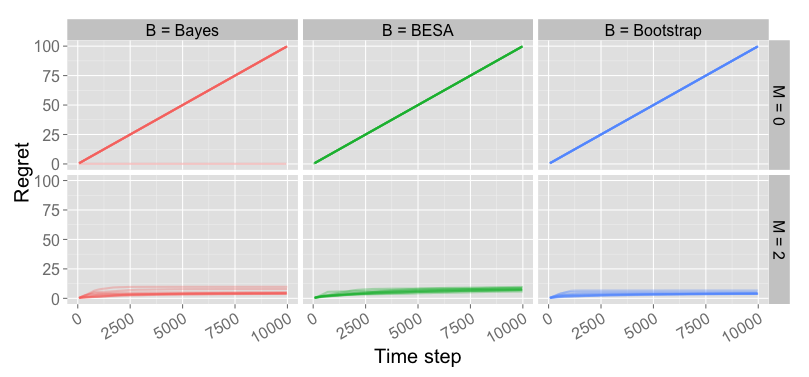}
    \caption{Cumulative regret of BootstrapThompson using different bootstrap methods (lower is better). Artificial prior data helps to drive efficient exploration.}
    \label{fig: regret}
\end{figure}

In Figure \ref{fig: regret} we plot the cumulative regret of three variants of BootstrapThompson using different bootstrap algorithms with $M=0$ or $M=2$.
For our simulations we set $\epsilon = 0.01$ and ran $20$ Monte Carlo simulations for each variant of the algorithm.  In each simulation and each time
period, the pair of artificial data points for cases with $M=2$ was sampled from a distribution $\tilde{P}$ that selects each of the two actions once and
samples an observation uniformly from $[0,1]$ for each.
We found that, in this example, the choice of bootstrap method makes little difference but that injecting
artificial data is crucial to incentivizing efficient exploration.

The six subplots in Figure \ref{fig: regret} present results differentiated by choice of bootstrap algorithms and whether or not to include artificial data.
The columns indicate which bootstrap algorithm is used with labels ``Bootstrap'' for Algorithm \ref{alg:Bootstrap}, ``Bayes'' for Algorithm \ref{alg:BayesBootstrap}, and ``BESA'' for a recently proposed bootstrap approach.
The rows indicate whether or not artificial prior data was used.
We see that BootstrapThompson generally fails to learn with $M=0$, however the inclusion of artificial data helps to drive efficient exploration.
The choice of bootstrap algorithm $B$ seems to make relatively little difference, however we do find that Algorithms \ref{alg:Bootstrap} and \ref{alg:BayesBootstrap} seem to outperform BESA on this example\footnote{The BESA algorithm is a variant of the bootstrap that applies to two armed bandit problems. In each time period, the algorithm estimates the reward of each arm by drawing a sample average (with replacement) with sample size equal to the number of times the \emph{other} arm has been played.  This apparently performs well in some settings \cite{baransi2014sub}, but the approach does not generalize gracefully to settings with dependent arms.}.

\subsection{Analysis}
\label{sec: Regret}

The BootstrapThompson algorithm is similar to Thompson sampling, the only difference being that a draw from the posterior distribution is
replaced by a bootstrap sample.
In fact, we can show that, for particular choices of bootstrap algorithm and artificial data distribution, the two algorithms are equivalent.

Consider as an example a multi-armed bandit problem with independent arms, for which each $a$th arm generates rewards from a Bernoulli distribution
with mean $\theta_a$.  Suppose that our prior distribution over $\theta_a$ is $\text{Beta}(\alpha_a,\beta_a)$.  Then, the posterior conditioned on
observing $n_{a0}$ outcomes with reward zero and $n_{a1}$ outcomes with reward one is $\text{Beta}(\alpha_a + n_{a0},\beta_a + n_{a1})$.
If $\alpha_a$ and $\beta_a$ are positive integers, a sample $\hat{\theta}_a$ from this distribution can be generated by the following procedure:
sample $x_1,\ldots,x_{\alpha_a+n_{a1}}, y_1,\ldots, y_{\beta_a+n_{a2}} \sim \text{Exp}(1)$ and let $\hat{\theta}_a = \sum_i x_i / (\sum_i x_i + \sum_j y_j)$.
This sampling procedure is identical to BayesBootstrap (Algorithm \ref{alg:BayesBootstrap}) with artificial data generated by a distribution $\tilde{P}$
that assigns all probability to a single outcome that, for each arm $a$, produces $\alpha_a+\beta_a$ data samples,
with $\alpha_a$ of them associated with reward one and $\beta_a$ of them associated with reward $0$.

The example we have presented can easily be generalized to the case where each arm generates rewards from among a finite set of possibilities
with probabilities distributed according to a Dirichlet prior.  We expect that, with appropriately designed schemes for generating artificial data,
such equivalences can also be established for a far broader range of problems.

The aforementioned equivalencies imply that theoretical regret bounds previously developed for Thompson sampling \cite{russo2014,agrawal2012further}
apply to the BootstrapThompson algorithm with the Bayesian bootstrap and appropriately generated artificial data.



\section{Reinforcement learning}

In reinforcement learning, actions taken by the agent can impose delayed consequences.  This makes the design of exploration strategies more
challenging than for multi-armed bandit problems.  To fix a context, consider an agent that interacts with an environment over repeated episodes of length $\tau$.
In each time period $t=1,..,\tau$ of each episode episode $l=1,2,..$, the agent observes a state $s_{lt}$ and selects an action $a_{lt}$ according to a
policy $\pi$ which maps states to actions.  A reward of $r_{lt}$ and state transition to $s_{l t+1}$ are then realized.
The agent's goal is to maximize the long term sum of expected rewards, even though she is initially unsure of the system dynamics and reward structure.

A common approach to reinforcement learning involves learning a state-action value function $Q$, which for each time $t$, state $s$, and action $a$,
provides an estimate $Q_t(s,a)$ of expected rewards over the remainder of the episode: $r_{lt} + r_{l,t+1} + \cdots + r_{l,\tau}$.  Given a state-action value function $Q$,
it is natural for the agent to select an action that maximizes $Q_t(s,a)$ when at state $s$ at time $t$.

There is a large literature on reinforcement learning algorithms which balance exploration with exploitation in a variety of ways \cite{brafman2003r,jaksch2010near,osband2013more}.
However, the vast majority of these algorithms operate in the ``tabula rasa'' setting, which does not allow for generalization between state-action pairs.
For most practical systems where the numbers of states and actions is very large or even infinite the ability to generalize is crucial for good performance.
Of those algorithms which do combine generalization with exploration, many require an intractable model-based planning step, or are restricted to unrealistic parametric domains \cite{osband2014near,osband2014model,abbasi2011regret}.

By contrast, some of the most successful applications of reinforcement learning generalize using nonlinearly parameterized models, like deep
neural networks, that approximate the state-action value function \cite{tesauro1995temporal,mnih2015human}.
These algorithms have attained superhuman performance and generated excitement for a new wave of artificial intelligence, but still fail at simple tasks that require efficient
exploration since they use simple exploration schemes that do not adequately account for the possibility of delayed consequences.
Recent research has shown how to combine efficient generalization and exploration via randomized linearly parameterized value functions \cite{van2014generalization}.
The approach presented in \cite{van2014generalization} can be viewed as a variant of Thompson sampling applied in a reinforcement learning context.  But this approach \cite{van2014generalization} does not serve the needs of nonlinear parameterizations.  What we present now, as Algorithm \ref{alg: deeper}, is a version of Thompson
sampling that does serve such needs via leveraging the bootstrap and artificial data.

\begin{algorithm}[H]
\caption{\protect\\ Reinforcement Learning with Bootstrapped Value Function Randomization}
\label{alg: deeper}
{\small
\begin{algorithmic}[1]
\State{\textbf{Input:} Bootstrap algorithm $B$, value function approximator $\phi$, \\
\hspace{11.5mm} number of artificial episodes $M$, sampling distribution $\tilde{P}$}
\State{$H = ()$}
\For{episode $l=1,2,..$}
\State{Sample $\tilde{H} = ((\tilde{s}_{11}, \tilde{a}_{11}, \tilde{r}_{11}, \ldots \tilde{s}_{1\tau}, \tilde{a}_{l\tau}, \tilde{r}_{l\tau}), \ldots,
(\tilde{s}_{M1}, \tilde{a}_{M1}, \tilde{r}_{M1}, \ldots \tilde{s}_{M\tau}, \tilde{a}_{M \tau}, \tilde{r}_{M \tau})) \sim \tilde{P}$}
\State{Bootstrap sample $\hat{P} \leftarrow \text{B}(\tilde{H} \cup H, \ \phi, \ K=1)$}
\State{Sample $Q \sim \hat{P}$}
\For{time $t=1,..,\tau$}
    \State{Select action $a_{lt} \in \arg \max_\alpha Q_t(s_{lt},\alpha)$}
    \State{Observe reward $r_{lt}$, transition to $s_{l,t+1}$}
\EndFor
    \State{Update $H \leftarrow H \cup (s_{l1}, a_{l1}, r_{l1}, \ldots, s_{l\tau}, a_{l\tau},r_{l\tau})$}
\EndFor
\end{algorithmic}
}
\end{algorithm}

In the context of our episodic setting, each element of the data set corresponds to a sequence of observations made over an episode.
The algorithm takes as input a function $\phi$, which should itself be viewed as an algorithm that estimates the state-action value function from
this data set.  For example, $\phi$ could output a deep neural network trained to fit a state-action value function via least-squares
value iteration.
A number of conventional reinforcement learning algorithms would fit the state-action value function to the observed history $H$.
Two key features that distinguishes Algorithm \ref{alg: deeper} is that the state-action value function is fit to a random subsample of
data and that this subsample is drawn from a combination of historical and artificial data.

Before the beginning of each episode, the algorithm applies $\phi$ to generate a randomized state-action value function.
The agent then follows the greedy policy with respect to that sample over the entire episode.
As is more broadly the case with Thompson sampling, the algorithm balances exploration with exploitation through the randomness of these samples.
The algorithm enjoys the benefits of what we call {\it deep exploration} in that it sometimes selects actions which are neither exploitative
nor informative in themselves, but that are oriented toward positioning the agent to gain useful information downstream in the episode.
In fact, the general approach represented by this algorithm may be the only known computationally efficient means of achieving deep exploration
with nonlinearly parameterized representations such as deep neural networks.

As discussed earlier, the inclusion of an artificial history can be crucial to incentivize proper exploration in multi-armed bandit problems.
The same is true for reinforcement learning.  One simple approach to generating artificial data that accomplishes this
in the context of Algorithm \ref{alg: deeper} is to sample state-action pairs from a diffusely mixed
generative model and assign them stochastically optimistic rewards (see \cite{van2014generalization} for a definition) and random state transitions.
When prior data is available from episodes of experience with actions selected by an expert agent, one can also augment this artificial data with that history
of experience.  This offers a means of incorporating apprenticeship learning as a springboard for the learning process.

Fitting a model like a deep neural network can itself be a computationally expensive task.  As such it is desirable to use incremental methods
that incorporate new data samples into the fitting process as they appear, without having to refit the model from scratch.
It is important to observe that a slight variation of Algorithm \ref{alg: deeper} accommodates this sort of incremental fitting by leveraging parallel computation.
This variation is presented as Algorithms \ref{alg:BayesBootstrap2} and \ref{alg: deeper2}.  The algorithm makes use of an incremental model
learning method $\phi$, which takes as input a current model, previous data set, and new data point, with a weight assigned to each data point.
Algorithm \ref{alg: deeper2} maintains $K$ models (for example, $K$ deep
neural networks), incrementally updating each in parallel after observing each episode.  The model used to guide action in an episode is
sampled uniformly from the set of $K$.  It is worth noting that this is akin to training each model using experience replay, but with past experiences
weighted randomly to induce exploration.

\begin{algorithm}[H]
\caption{\protect\\ IncrementalBayesBootstrapSample}
\label{alg:BayesBootstrap2}
{\textbf{Input:} Data $x_1,..,x_N \in \mathcal{X}$, weights $w_1,\ldots,w_{N-1} \in \Re_+$,
function $\phi: \mathcal{P}(\mathcal{X}) \mapsto \mathcal{Y}$} \\
{\textbf{Output:} Sampled weight $w_N$, sampled outcome $y$}
{\small
\begin{algorithmic}[1]
    \State{sample $w_N \sim {\rm Exp}(1)$}
    \State{for all $dx \subseteq \mathcal{X}$, let $\hat{P}(dx) = \sum_{n=1}^N w_n {\bf 1}(x_n \in dx) / \sum_{n=1}^N w_n$}
    \State{compute $y = \phi(\hat{P})$}
\end{algorithmic}
}
\end{algorithm}
\vspace{-3mm}

\begin{algorithm}[H]
\caption{\protect\\ Incremental Reinforcement Learning with Bootstrapped Value Function Randomization}
\label{alg: deeper2}
{\small
\begin{algorithmic}[1]
\algblock{Parfor}{EndParfor}
\State{\textbf{Input:} Incremental bootstrap algorithm $B$, value function approximator $\phi$, \\
\hspace{11.5mm} number of models $K$, initial models $Q^1,\ldots,Q^K$ \\
\hspace{11.5mm} number of artificial episodes $M$, sampling distribution $\tilde{P}$}
\Parfor{ bootstrap sample $k=1,\ldots,K$}
\State{$H^k = ()$}
\State{Sample $\tilde{H}^k = ((\tilde{s}^k_{11}, \tilde{a}^k_{11}, \tilde{r}^k_{11}, \ldots \tilde{s}^k_{1\tau}, \tilde{a}^k_{l\tau}, \tilde{r}^k_{l\tau}), \ldots,
(\tilde{s}^k_{M1}, \tilde{a}^k_{M1}, \tilde{r}^k_{M1}, \ldots \tilde{s}^k_{M\tau}, \tilde{a}^k_{M \tau}, \tilde{r}^k_{M \tau})) \sim \tilde{P}$}
\EndParfor
\For{episode $l=1,2,..$}
\Parfor{ bootstrap sample $k=1,\ldots,K$}
\State{Bootstrap sample $Q^k \leftarrow \text{B}(\tilde{H}^k \cup \tilde{H}^k, \ \phi, Q^k)$}
\EndParfor
\State{Sample $k \sim \text{unif}(1,\ldots,K)$}
\For{time $t=1,..,\tau$}
    \State{Select action $a_{lt} \in \arg \max_\alpha Q_t^k(s_{lt},\alpha)$}
    \State{Observe reward $r_{lt}$, transition to $s_{l,t+1}$}
\EndFor
\Parfor{ bootstrap sample $k=1,\ldots,K$}
    \State{Update $H^k \leftarrow H^k \cup (s_{l1}, a_{l1}, r_{l1}, \ldots, s_{l\tau}, a_{l\tau},r_{l\tau})$}
\EndParfor
\EndFor
\end{algorithmic}
}
\end{algorithm}

\newpage
\small{
\bibliography{referenceInformation.bib}
\bibliographystyle{unsrt}
}

\end{document}